# Wrapper Maintenance: A Machine Learning Approach

**Kristina Lerman**　　　　　　　　　　　　　　　　　　LERMAN@ISI.EDU
*USC Information Sciences Institute*
*4676 Admiralty Way*
*Marina del Rey, CA 90292 USA*

**Steven N. Minton**　　　　　　　　　　　　　　　　　MINTON@FETCH.COM
*Fetch Technologies*
*4676 Admiralty Way*
*Marina del Rey, CA 90292 USA*

**Craig A. Knoblock**　　　　　　　　　　　　　　　　KNOBLOCK@ISI.EDU
*USC Information Sciences Institute and Fetch Technologies*
*4676 Admiralty Way*
*Marina del Rey, CA 90292 USA*

## Abstract

The proliferation of online information sources has led to an increased use of wrappers for extracting data from Web sources. While most of the previous research has focused on quick and efficient generation of wrappers, the development of tools for wrapper maintenance has received less attention. This is an important research problem because Web sources often change in ways that prevent the wrappers from extracting data correctly. We present an efficient algorithm that learns structural information about data from positive examples alone. We describe how this information can be used for two wrapper maintenance applications: wrapper verification and reinduction. The wrapper verification system detects when a wrapper is not extracting correct data, usually because the Web source has changed its format. The reinduction algorithm automatically recovers from changes in the Web source by identifying data on Web pages so that a new wrapper may be generated for this source. To validate our approach, we monitored 27 wrappers over a period of a year. The verification algorithm correctly discovered 35 of the 37 wrapper changes, and made 16 mistakes, resulting in precision of 0.73 and recall of 0.95. We validated the reinduction algorithm on ten Web sources. We were able to successfully reinduce the wrappers, obtaining precision and recall values of 0.90 and 0.80 on the data extraction task.

## 1. Introduction

There is a tremendous amount of information available online, but much of this information is formatted to be easily read by human users, not computer applications. Extracting information from semi-structured Web pages is an increasingly important capability for Web-based software applications that perform information management functions, such as shopping agents (Doorenbos, Etzioni, & Weld, 1997) and virtual travel assistants (Knoblock, Minton, Ambite, Muslea, Oh, & Frank, 2001b; Ambite, Barish, Knoblock, Muslea, Oh, & Minton, 2002), among others. These applications, often referred to as agents, rely on Web wrappers that extract information from semi-structured sources and convert it to a structured format. Semi-structured sources are those that have no explicitly specified grammar or schema, but have an implicit grammar that can be used to identify relevant information on the page. Even text sources such as email messages have some structure in the heading that can be exploited to extract the date, sender, addressee, title, and body of the messages. Other sources, such as online catalogs, have a very regular structure that can be exploited to extract all the data automatically.

Wrappers rely on extraction rules to identify the data field to be extracted. Semi-automatic creation of extraction rules, or wrapper induction, has been an active area of research in recent years (Knoblock, Lerman, Minton, & Muslea, 2001a; Kushmerick, Weld, & Doorenbos, 1997). The most advanced of these wrapper generation systems use machine learning techniques to learn the extraction rules by example. For instance, the wrapper induction tool developed at USC (Knoblock et al., 2001a; Muslea, Minton, & Knoblock, 1998) and commercialized by Fetch Technologies, allows the user to mark up data to be extracted on several example pages from an online source using a graphical user interface. The system then generates "landmark"-based extraction rules for these data that rely on the page layout. The USC wrapper tool is able to efficiently create extraction rules from a small number of examples; moreover, it can extract data from pages that contain lists, nested structures, and other complicated formatting layouts.

In comparison to wrapper induction, wrapper maintenance has received less attention. This is an important problem, because even slight changes in the Web page layout can break a wrapper that uses landmark-based rules and prevent it from extracting data correctly. In this paper we discuss our approach to the wrapper maintenance problem, which consists of two parts: wrapper verification and reinduction. A *wrapper verification* system monitors the validity of data returned by the wrapper. If the site changes, the wrapper may extract nothing at all or some data that is not correct. The verification system will detect data inconsistency and notify the operator or automatically launch a wrapper repair process. A *wrapper reinduction* system repairs the extraction rules so that the wrapper works on changed pages.

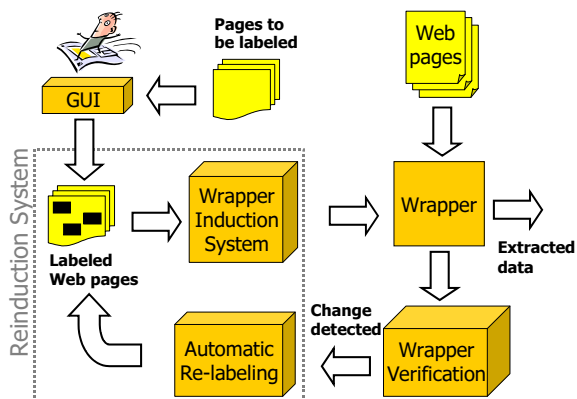

Figure 1: Life cycle of a wrapper

Figure 1 graphically illustrates the entire life cycle of a wrapper. As shown in the figure, the wrapper induction system takes a set of web pages labeled with examples of the data to be extracted. The output of the wrapper induction system is a wrapper, consisting of a set of extraction rules that describe how to locate the desired information on a Web page. The wrapper verification system uses the functioning wrapper to collect extracted data. It then learns patterns describing the structure of data. These patterns are used to verify that the wrapper is correctly extracting data at a later date. If a change is detected, the system can automatically repair a wrapper by using this structural information to locate examples of data on the new pages and re-running the wrapper induction system with these examples. At the core of these wrapper maintenance applications is a machine learning algorithm that learns structural information about common data fields. In this paper we introduce the algorithm, DATAPROG, and describe its application to the wrapper maintenance tasks in detail. Though we focus on web applications, the learning technique is not web-specific, and can be used for data validation in general.

Note that we distinguish two types of extraction rules: landmark-based rules that extract data by exploiting the structure of the Web page, and content-based rules, which we refer to as content patterns or simply patterns, that exploit the structure of the field itself. Our previous work focused on learning landmark rules for information extraction (Muslea, Minton, & Knoblock, 2001). The current work shows that augmenting these rules with content-based patterns provides a foundation for sophisticated wrapper maintenance applications.

## 2. Learning Content Patterns

The goal of our research is to extract information from semi-structured information sources. This typically involves identifying small chunks of highly informative data on formatted pages (as opposed to parsing natural language text). Either by convention or design, these *fields* are usually structured: phone numbers, prices, dates, street addresses, names, schedules, *etc.* Several examples of street addresses are given in Fig. 2. Clearly, these strings are not arbitrary, but share some similarities. The objective of our work is to learn the structure of such fields.

<div align="center">

4676 Admiralty Way
10924 Pico Boulevard
512 Oak Street
2431 Main Street
5257 Adams Boulevard

</div>

Figure 2: Examples of a street address field

### 2.1 Data Representation

In previous work, researchers described the fields extracted from Web pages by a character-level grammar (Goan, Benson, & Etzioni, 1996) or a collection of global features, such as the number of words and the density of numeric characters (Kushmerick, 1999). We employ an intermediate word-level representation that balances the descriptive power and specificity of the character-level representation with the compactness and computational efficiency of the global representation. Words, or more accurately tokens, are strings generated from an alphabet containing different types of characters: alphabetic, numeric, punctuation, *etc.* We use the token's character types to assign it to one or more syntactic categories: alphabetic, numeric, *etc.* These categories form a hierarchy depicted in Fig. 3, where the arrows point from more general to less general categories. A unique specific token type is created for every string that appears in at least $k$ examples, as determined in a preprocessing step. The hierarchical representation allows for multi-level generalization. Thus, the token "Boulevard" belongs to the general token types Alphanum (alphanumeric strings), Alpha (alphabetic strings), Upper (capitalized words), as well as to the specific type representing the string "Boulevard". This representation is flexible and may be expanded to include domain specific information. For example, the numeric type is divided into categories that include range information about the number — Large (larger than 1000), Medium (medium numbers, between 10 and 1000) and Small (smaller than 10)— and number of digits: $1-$, $2-$, and $3-$digit. Likewise, we may explicitly include knowledge about the type of information being parsed, *e.g.*, some 5-digit numbers could be represented as zipcode.

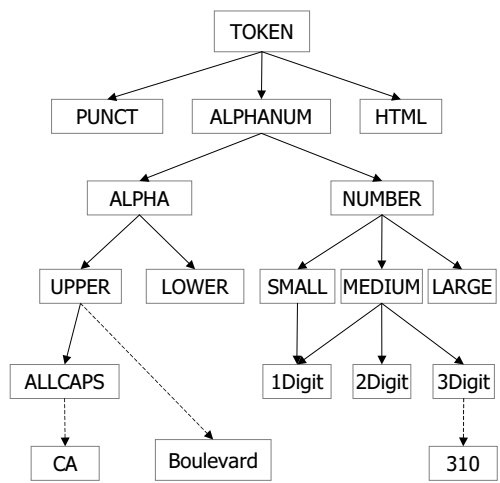

Figure 3: Portion of the token type syntactic hierarchy

We have found that a sequence of specific and general token types is more useful for describing the content of information than the character-level finite state representations used in previous work (Carrasco & Oncina, 1994; Goan et al., 1996). The character-level description is far too fine grained to compactly describe data and, therefore, leads to poor generality. The coarse-grained token-level representation is more appropriate for most Web data types. In addition, the data representation schemes used in previous work attempt to describe the entire data field, while we use only the starting and ending sequences, or patterns, of tokens to capture the structure of the data fields. The reason for this is similar to the one above: using the starting and ending patterns allows us to generalize the structural information for many complex fields which have a lot of variability. Such fields, *e.g.*, addresses, usually have some regularity in how they start and end that we can exploit. We call the starting and ending patterns collectively a *data prototype*. As an example, consider a set of street addresses in Fig. 2. All of the examples start with a pattern <Number Upper> and end with a specific type <Boulevard> or more generally <Upper>. Note that the pattern language does not allow loops or recursion. We believe that recursive expressions are not useful representations of the types of data we are trying to learn, because they are harder to learn and lead to over-generalization.

## 2.2 Learning from Positive Examples

The problem of learning the data prototype from a set of examples that are labeled as belonging (or not) to a class may be stated in one of two related ways: as a classification or as a conservation task. In the classification task, both positive and negative instances of the class are used to learn a rule that will correctly classify new examples. Classification algorithms, like FOIL (Quinlan, 1990), use negative examples to guide the specialization of the rule. They construct discriminating descriptions — those that are satisfied by the positive examples and not the negative examples. The conservation task, on the other hand, attempts to find a characteristic description (Dietterich & Michalski, 1981) or conserved patterns (Brazma, Jonassen, Eidhammer, & Gilbert, 1995), in a set of positive examples of a class. Unlike the discriminating description, the characteristic description will often include redundant features. For example, when learning a description of street addresses, with city names serving as negative examples, a classification algorithm will learn that <Number> is a good description, because all the street addresses start with it and none of the city names do. The capitalized word that follows the number in addresses is a redundant feature, because it does not add to the discriminating power of the learned description. However, if an application using this description encounters a zipcode in the future, it will incorrectly classify it as a street address. This problem could have been avoided if <Number Upper> was learned as a description of street addresses. Therefore, when negative examples are not available to the learning algorithm, the description has to capture all the regularity of data, including the redundant features, in order to correctly identify new instances of the class and differentiate them from other classes. Ideally, the characteristic description learned from positive examples alone is the same as the discriminating description learned by the classification algorithm from positive and negative examples, where negative examples are drawn from infinitely many classes. While most of the widely used machine learning algorithms (*e.g.*, decision trees (Quinlan, 1993), inductive logic programming (Muggleton, 1991)) solve the classification task, there are fewer algorithms that learn characteristic descriptions.

In our applications, an appropriate source of negative examples is problematic; therefore, we chose to frame the learning problem as a conservation task. We introduce an algorithm that learns data prototypes from positive examples of the data field alone. The algorithm finds statistically significant sequences of tokens. A sequence of token types is significant if it occurs more frequently than would be expected if the tokens were generated randomly and independently of one another. In other words, each such sequence constitutes a pattern that describes many of the positive examples of data and is highly unlikely to have been generated by chance.

The algorithm estimates the baseline probability of a token type's occurrence from the proportion of all types in the examples of the data field that are of that type. Suppose we are learning a description of the set of street addresses in Fig. 2, and have already found a significant token sequence — *e.g.*, the pattern consisting of the single token <Number> — and want to determine whether the more specific pattern, <Number Upper>, is also a significant pattern. Knowing the probability of occurrence of the type Upper, we can compute how many times Upper can be expected to follow Number completely by chance. If we observe a considerably greater number of these sequences, we conclude that the longer pattern is also significant.

We use hypothesis testing (Papoulis, 1990) to decide whether a pattern is significant. The null hypothesis is that observed instances of this pattern were generated by chance, via the random, independent generation of the individual token types. Hypothesis testing decides, at a given confidence level, whether the data supports rejecting the null hypothesis. Suppose $n$ identical sequences have been generated by a random source. The probability that a token type $T$ (whose overall probability of occurrence is $p$) will be the next type in $k$ of these sequences has a binomial distribution. For a large $n$, the binomial distribution approaches a normal distribution $P(x, \mu, \sigma)$ with $\mu = np$ and $\sigma^2 = np(1-p)$. The cumulative probability is the probability of observing at least $n_1$ events:

$$P(k \geq n_1) = \int_{n_1}^{\infty} P(x, \mu, \sigma) dx \tag{1}$$

We use polynomial approximation formulas (Abramowitz & Stegun, 1964) to compute the value of the integral.

The significance level of the test, $\alpha$, is the probability that the null hypothesis is rejected even though it is true, and it is given by the cumulative probability above. Suppose we set $\alpha = 0.05$. This means that we expect to observe at least $n_1$ events 5% of the time under the null hypothesis. If the number of observed events is greater, we reject the null hypothesis (at the given significance level), *i.e.*, decide that the observation is significant. Note that the hypothesis we test is derived from observation (data). This constraint reduces the number of degrees of freedom of the test; therefore, we must subtract one from the number of observed events. This also prevents the anomalous case when a single occurrence of a rare event is judged to be significant.

### 2.3 DataProG Algorithm

We now describe DataProG, the algorithm that finds statistically significant patterns in a set of token sequences. During the preprocessing step the text is tokenized, and the tokens are assigned one or more syntactic types (see Figure 3). The patterns are encoded in a type of prefix tree, where each node corresponds to a token type. DataProG relies on significance judgements to grow the tree and prune the nodes. Every path through the resulting tree starting at the root node corresponds to a significant pattern found by the algorithm. In this section, we focus the discussion on the version of the algorithm that learns starting patterns. The algorithm is easily adapted to learn ending patterns.

We present the pseudocode of the DataProG algorithm in Table 1. DataProG grows the pattern tree incrementally by (1) finding all significant specializations (*i.e.*, longer patterns) of a pattern and (2) pruning the less significant of the generalizations (or specializations) among patterns of the same length. As the last step, DataProG extracts all significant patterns from the pattern tree, including those generalizations (*i.e.*, shorter patterns) found to be significant given the more specific (*i.e.*, longer) patterns.

DATAPROG MAIN LOOP
Create root node of tree;
For next node Q of tree
       Create children of Q;
       Prune nodes;
Extract patterns from tree;

CREATE CHILDREN OF Q
For each token type T at next position in examples
    Let C = NewNode;
    Let $C.token$ = T;
    Let $C.examples$ = Q.examples that are followed by T;
    Let $C.count$ = $|C.examples|$;
    Let $C.pattern$ = concat($Q.pattern\,T$);
    If Significant($C.count$, $Q.count$, $T.probability$)
               AddChildToTree(C, Q);
    End If
End T loop

PRUNE NODES
For each child C of Q
    For each sibling S of C s.t. $S.pattern \subset C.pattern$
        Let $N = C.count - S.count$
        If Not(Significant($N, Q.count, C.token.probability$))
            Delete C;
            break;
        Else
            Delete S;
        End If
    End S loop
End C loop

EXTRACT PATTERNS FROM TREE
Create empty list;
For every node Q of tree
    For every child C of Q
        Let $N = C.count - \sum_i(S_i.count | S_i \in Children(C))$
        If Significant($N, Q.count, C.token.probability$)
           Add $C.pattern$ to the list;
Return (list of patterns);

Table 1: Pseudocode of the DataProg algorithm

The tree is empty initially, and children are added to the root node. The children represent all tokens that occur in the first position in the training examples more often than expected by chance. For example, when learning addresses from the examples in Fig. 2, the root will have two child nodes: Alphanum and Number. The tree is extended incrementally at each node $Q$. A new child is added to $Q$ for every significant specialization of the pattern ending at $Q$. As explained previously, a child node is judged to be significant with respect to its parent node if the number of occurrences of the pattern ending at the child node is sufficiently large, given the number of occurrences of the pattern ending at the parent node and the baseline probability of the token type used to extend the pattern. To illustrate on our addresses example, suppose we have already found that a pattern <Number Upper> is significant. There are five ways to extend the tree (see Fig. 4) given the data: <Number Upper Alphanum>, <Number Upper Alpha>, <Number Upper Upper>, <Number Upper Street>, <Number Upper Boulevard>, and <Number Upper Way>. All but the last of these patterns are judged to be significant at $\alpha = 0.05$. For example, <Number Upper Upper> is significant, because Upper follows the pattern <Number Upper> five out of five times,[1] and the probability of observing at least that many longer sequences purely by chance is 0.0002.[2] Since this probability is less than $\alpha$, we judge this sequence to be significant.

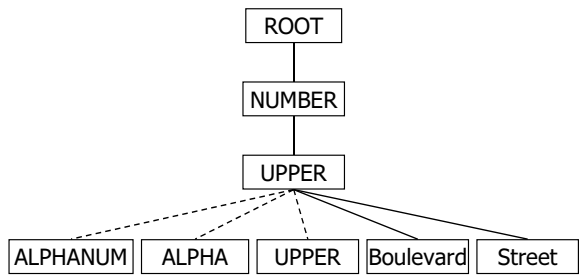

Figure 4: Pattern tree that describes the structure of addresses. Dashed lines link to nodes that are deleted during the pruning step.

The next step is to prune the tree. The algorithm examines each pair of sibling nodes, one of which is more general than the other, and eliminates the less significant of the pair. More precisely, the algorithm iterates through the newly created children of $Q$, from the most to least general, and for every pair of children $C_i$ and $C_j$, such that $C_i.pattern \subset C_j.pattern$ (i.e., $C_j.pattern$ is strictly more general than $C_i.pattern$), the algorithm keeps only $C_j$ if it explains significantly more data; otherwise, it keeps only $C_i$.[3]

---

1. Such small numbers are used for illustrative purposes only — the typical data sets from which the patterns are learned are much larger.
2. The calculation of this cumulative probability depends on the occurrence probability of Upper. We count the occurrence of each token type independently of the others. In our example, occurrence probability (relative fraction) of type Upper is 0.18.
3. DataProg is based on an earlier version of the algorithm, DataPro, described in the conference paper (Lerman & Minton, 2000). Note that in the original version of the algorithm, the specific patterns were always kept, regardless of whether the more general patterns were found to be significant or not.

Let us illustrate the pruning step with the example pattern tree in Fig. 4. We can eliminate the node Alpha­Num, because all the examples that match the pattern <Number Upper Alphanum> also match the pattern <Number Upper Alpha> — thus, Alphanum is not significant given its specialization Alpha. We can eliminate node Alpha for a similar reason. Next, we check whether <Number Upper Upper> is significant given the patterns <Number Upper Boulevard> and <Number Upper Street>. There are 2 instances of the address field that match the pattern <Number Upper Boulevard>, and 2 addresses that match <Number Upper Street>. If <Number Upper Upper> matches significantly more than 4 addresses, it will be retained and the more specific patterns will be pruned from the tree; otherwise, it will be deleted and the more specific ones kept. Because every example is described by at most one pattern of a given length, the pruning step ensures that the size of the tree remains polynomial in the number of tokens, thereby, guaranteeing a reasonable performance of the algorithm.

Once the entire tree has been expanded, the final step is to extract all significant patterns from the tree. Here, the algorithm judges whether the shorter (more general) pattern, *e.g.*, <Number Upper>, is significant given the longer specializations of it, *e.g.*, <Number Upper Boulevard> and <Number Upper Street>. This amounts to testing whether the excess number of examples that are explained by the shorter pattern, and not by the longer patterns, is significant. Any pattern that ends at a terminal node of the tree is significant. Note that the set of significant patterns may not cover all the examples in the data set, just a fraction of them that occur more frequently than expected by chance (at some significance level). Tables 2–4 show examples of several data fields from a yellow pages source (*Bigbook*) and a stock quote source (*Yahoo Quote*), as well as the starting patterns learned for each field.

## 3. Applications of Pattern Learning

As we explained in the introduction, wrapper induction systems use information from the layout of Web pages to create data extraction rules and are therefore vulnerable to changes in the layout, which occur frequently when the site is redesigned. In some cases the wrapper continues to extract, but the data is no longer correct. The output of the wrapper may also change because the format of the source data itself has changed: *e.g.*, when "$" is dropped from the price field ("9.95" instead of "$9.95"), or book availability changes from "Ships immediately" to "In Stock: ships immediately." Because other applications, such as Web agents (Ambite et al., 2002; Chalupsky et al., 2001), rely on data extracted by wrappers, wrapper maintenance is an important research problem. We divide the wrapper maintenance problem into two parts, each described separately in the paper. *Wrapper verification* automatically detects when a wrapper is not extracting data correctly from a Web source, while *wrapper reinduction* automatically fixes broken wrappers. Both applications learn a description of data, of which patterns learned by DataProg are a significant part.

---

This introduced a strong bias for specific patterns into the results, which led to a high proportion of false positives during the wrapper verification experiments. Eliminating the specificity bias, improved the performance of the algorithm on the verification task.

| BUSINESS NAME | ADDRESS |
|---|---|
| Chado Tea House | 8422 West 1st Street |
| Saladang | 363 South Fair Oaks Avenue |
| Information Sciences Institute | 4676 Admiralty Way |
| Chaya Venice | 110 Navy Street |
| Acorda Therapeutics | 330 West 58th Street |
| Cajun Kitchen | 420 South Fairview Avenue |
| Advanced Medical Billing Services | 9478 River Road |
| Vega 1 Electrical Corporation | 1723 East 8th Street |
| 21st Century Foundation | 100 East 85th Street |
| TIS the Season Gift Shop | 15 Lincoln Road |
| Hide Sushi Japanese Restaurant | 2040 Sawtelle Boulevard |
| Afloat Sushi | 87 East Colorado Boulevard |
| Prebica Coffee & Cafe | 4325 Glencoe Avenue |
| L ' Orangerie | 903 North La Cienega Boulevard |
| Emils Hardware | 2525 South Robertson Boulevard |
| Natalee Thai Restaurant | 998 South Robertson Boulevard |
| Casablanca | 220 Lincoln Boulevard |
| Antica Pizzeria | 13455 Maxella Avenue |
| NOBU Photographic Studio | 236 West 27th Street |
| Lotus Eaters | 182 5th Avenue |
| Essex On Coney | 1359 Coney Island Avenue |
| National Restaurant | 273 Brighton Beach Avenue |
| Siam Corner Cafe | 10438 National Boulevard |
| Grand Casino French Bakery | 3826 Main Street |
| Alejo ' s Presto Trattoria | 4002 Lincoln Boulevard |
| Titos Tacos Mexican Restaurant Inc | 11222 Washington Place |
| Killer Shrimp | 523 Washington Boulevard |
| Manhattan Wonton CO | 8475 Melrose Place |
| Starting patterns | |
| <Alpha Upper> | <Number Upper Upper> |
| <Alpha Upper Upper Restaurant> | <Number Upper Upper Avenue> |
| <Alpha '> | <Number Upper Upper Boulevard> |

Table 2: Examples of the business name and address fields from the *Bigbook* source, and the patterns learned from them

| CITY | STATE | PHONE |
|---|---|---|
| Los Angeles | CA | ( 323 ) 655 - 2056 |
| Pasadena | CA | ( 626 ) 793 - 8123 |
| Marina Del Rey | CA | ( 310 ) 822 - 1511 |
| Venice | CA | ( 310 ) 396 - 1179 |
| New York | NY | ( 212 ) 376 - 7552 |
| Goleta | CA | ( 805 ) 683 - 8864 |
| Marcy | NY | ( 315 ) 793 - 1871 |
| Brooklyn | NY | ( 718 ) 998 - 2550 |
| New York | NY | ( 212 ) 249 - 3612 |
| Buffalo | NY | ( 716 ) 839 - 5090 |
| Los Angeles | CA | ( 310 ) 477 - 7242 |
| Pasadena | CA | ( 626 ) 792 - 9779 |
| Marina Del Rey | CA | ( 310 ) 823 - 4446 |
| West Hollywood | CA | ( 310 ) 652 - 9770 |
| Los Angeles | CA | ( 310 ) 839 - 8571 |
| Los Angeles | CA | ( 310 ) 855 - 9380 |
| Venice | CA | ( 310 ) 392 - 5751 |
| Marina Del Rey | CA | ( 310 ) 577 - 8182 |
| New York | NY | ( 212 ) 924 - 7840 |
| New York | NY | ( 212 ) 929 - 4800 |
| Brooklyn | NY | ( 718 ) 253 - 1002 |
| Brooklyn | NY | ( 718 ) 646 - 1225 |
| Los Angeles | CA | ( 310 ) 559 - 1357 |
| Culver City | CA | ( 310 ) 202 - 6969 |
| Marina Del Rey | CA | ( 310 ) 822 - 0095 |
| Culver City | CA | ( 310 ) 391 - 5780 |
| Marina Del Rey | CA | ( 310 ) 578 - 2293 |
| West Hollywood | CA | ( 323 ) 655 - 6030 |
| Starting patterns | | |
| <Upper Upper> | <AllCaps> | <( 3digit ) 3digit - Large> |
| <Upper Upper Rey> | | |

Table 3: Examples of the city, state and phone number fields from the *Bigbook* source, and the patterns learned from them

| PRICE CHANGE | TICKER | VOLUME | PRICE |
|---|---|---|---|
| + 0 . 51 | INTC | 17 , 610 , 300 | 122 3 / 4 |
| + 1 . 51 | IBM | 4 , 922 , 400 | 109 5 / 16 |
| + 4 . 08 | AOL | 24 , 257 , 300 | 63 13 / 16 |
| + 0 . 83 | T | 8 , 504 , 000 | 53 1 / 16 |
| + 2 . 35 | LU | 9 , 789 , 300 | 68 |
| | ATHM | 5 , 646 , 400 | 29 7 / 8 |
| - 10 . 84 | COMS | 15 , 388 , 200 | 57 11 / 32 |
| - 1 . 24 | CSCO | 19 , 135 , 900 | 134 1 / 2 |
| - 1 . 59 | GTE | 1 , 414 , 900 | 65 15 / 16 |
| - 2 . 94 | AAPL | 2 , 291 , 800 | 117 3 / 4 |
| + 1 . 04 | MOT | 3 , 599 , 600 | 169 1 / 4 |
| - 0 . 81 | HWP | 2 , 147 , 700 | 145 5 / 16 |
| + 4 . 45 | DELL | 40 , 292 , 100 | 57 3 / 16 |
| + 0 . 16 | GM | 1 , 398 , 100 | 77 15 / 16 |
| - 3 . 48 | CIEN | 4 , 120 , 200 | 142 |
| + 0 . 49 | EGRP | 7 , 007 , 400 | 25 7 / 8 |
| - 3 . 38 | HLIT | 543 , 400 | 128 13 / 16 |
| + 1 . 15 | RIMM | 307 , 500 | 132 1 / 4 |
| | C | 6 , 145 , 400 | 49 15 / 16 |
| - 2 . 86 | GPS | 1 , 023 , 600 | 44 5 / 8 |
| - 6 . 46 | CFLO | 157 , 700 | 103 1 / 4 |
| - 0 . 82 | DCLK | 1 , 368 , 100 | 106 |
| + 2 . 00 | NT | 4 , 579 , 900 | 124 1 / 8 |
| + 0 . 13 | BFRE | 149 , 000 | 46 9 / 16 |
| - 1 . 63 | QCOM | 7 , 928 , 900 | 128 1 / 16 |
| Starting patterns | | | |
| <Punct 1digit . 2digit> | <AllCaps> | <Number , 3digit , 3digit> | <Medium 1digit / Number> <Medium 15 / 16 > |

Table 4: Data examples from the *Yahoo Quote* source, and the patterns learned from them

### 3.1 Wrapper Verification

If the data extracted by the wrapper changes significantly, this is an indication that the Web source may have changed its format. Our wrapper verification system uses examples of data extracted by the wrapper in the past that are known to be correct in order to acquire a description of the data. The learned description contains features of two types: patterns learned by DataProG and global numeric features, such as the density of tokens of a particular type. The application then checks that this description still applies to the new data extracted by the wrapper. Thus, wrapper verification is a specific instance of the data validation task.

The verification algorithm works in the following way. A set of queries is used to retrieve HTML pages from which the wrapper extracts (correct) training examples. The algorithm then computes the values of a vector of features, $\vec{k}$, that describes each field of the training examples. These features include the patterns that describe the common beginnings (or endings) of the field. During the verification phase, the wrapper generates a set of (new) test examples from pages retrieved using the same set of queries, and computes the feature vector $\vec{r}$ associated with each field of the test examples. If the two distributions, $\vec{k}$ and $\vec{r}$ (see Fig. 5), are statistically the same (at some significance level), the wrapper is judged to be extracting correctly; otherwise, it is judged to have failed.

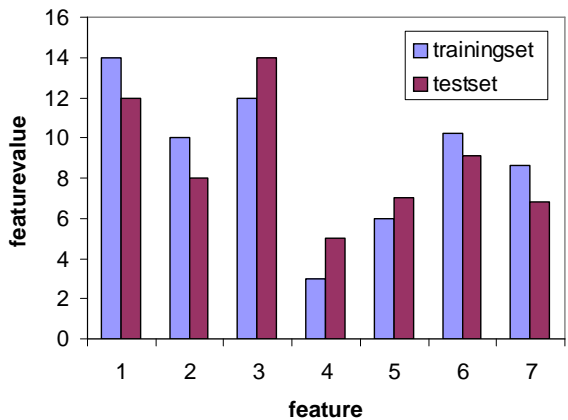

Figure 5: A hypothetical distribution of features over the training and test examples

Each field is described by a vector, whose $i$th component is the value of the $i$th feature, such as the number of examples that match pattern $j$. In addition to patterns, we use the following numeric features to describe the sets of training and test examples: the average number of tuples-per-page, mean number of tokens in the examples, mean token length, and the density of alphabetic, numeric, HTML-tag and punctuation types. We use goodness of fit method (Papoulis 1990) to decide whether the two distributions are the same. To use the goodness of fit method, we must first compute Pearson's test statistic for the data. The Pearson's test statistic is defined as:

$$q = \sum_{i=1}^{m} \frac{(t_i - e_i)^2}{e_i} \qquad (2)$$

where $t_i$ is the observed value of the $i$th feature in the test data, and $e_i$ is the expected value for that feature, and $m$ is the number of features. For the patterns $e_i = n r_i / N$, where $r_i$ is the number of training examples explained by the $i$th patter, $N$ is the number of examples in the training set and $n$ is the number of examples in the test set. For numeric features $e_i$ is simply the value of that feature for the training set. The test statistic $q$ has a chi-squared distribution with $m - 1$ independent degrees of freedom. If $q < \chi^2(m-1;\alpha)$, we conclude that at significance level $\alpha$ the two distributions are the same; otherwise, we conclude that they are different. Values of $\chi^2$ for different values of $\alpha$ and $m$ can be looked up in a statistics table or calculated using an approximation formula.

In order to use the test statistic reliably, it helps to use as many independent features as possible. In the series of verification experiments reported in (Lerman & Minton, 2000), we used the starting and ending patterns and the average number of tuples-per-page feature when computing the value of $q$. We found that this method tended to overestimate the test statistic, because the features (starting and ending patterns) were not independent. In the experiments reported in this paper, we use only the starting patterns, but in order to increase the number of features, we added numeric features to the description of data.

### 3.1.1 Results

We monitored 27 wrappers (representing 23 distinct Web sources) over a period of ten months, from May 1999 to March 2000. The sources are listed in Table 5. For each wrapper, the results of 15–30 queries were stored periodically, every 7–10 days. We used the same query set for each source, except for the *hotel* source, because it accepted dated queries, and we had to change the dates periodically to get valid results. Each set of new results (test examples) was compared with the last correct wrapper output (training examples).

The verification algorithm used DATAPROG to learn the starting patterns and numeric features for each field of the training examples and made a decision at a high significance level (corresponding to $\alpha = 0.001$) about whether the test set was statistically similar to the training set. If none of the starting patterns matched the test examples or if the data was found to have changed significantly for any data field, we concluded that the wrapper failed to extract correctly from the source; otherwise, if all the data fields returned statistically similar data, we concluded that the wrapper was working correctly.

A manual check of the 438 comparisons revealed 37 wrapper changes attributable to changes in the source layout and data format.[4] The verification algorithm correctly discovered 35 of these changes and made 15 mistakes. Of these mistakes, 13 were *false positives*, which means that the verification program decided that the wrapper failed when in reality it was working correctly. Only two of the errors were the more important *false negatives*, meaning that the algorithm did not detect a change in the data source. The numbers above

---

4. Seven of these were, in fact, internal to the wrapper itself, as when the wrapper was modified to extract "\$22.00" instead of "22.00" for the price field. Because these actions were mostly outside of our control, we chose to classify them as wrapper changes.

| Source | Type | Data Fields |
|--------|------|-------------|
| *airport* | tuple/list | airport code, name |
| *altavista* | list | url, title |
| *Amazon* | tuple | book author, title, price, availability, isbn |
| *arrowlist* | list | part number, manufacturer, price, status, description, url |
| *Bigbook* | tuple | business name, address, city, state, phone |
| *Barnes&Noble* | tuple | book author, title, price, availability, isbn |
| *borders* | list | book author, title, price, availability |
| *cuisinenet* | list | restaurant name, cuisine, address, city, state, phone, link |
| *geocoder* | tuple | latitude, longitude, street, city, state |
| *hotel* | list | name, price, distance, url |
| *mapquest* | tuple | hours, minutes, distance, url |
| *northernlight* | list | url, title |
| *parking* | list | lotname, dailyrate |
| *Quote* | tuple | stock ticker, price, pricechange, volume |
| *Smartpages* | tuple | name, address, city, state, phone |
| *showtimes* | list | movie, showtimes |
| *theatre* | list | theater name, url, address |
| *Washington Post* | tuple | taxi price |
| *whitepages* | list | business name, address, city, state, phone |
| *yahoo people* | list | name, address, city, state, phone |
| *Yahoo Quote* | tuple | stock ticker, price, pricechange, volume |
| *yahoo weather* | tuple | temperature, forecast |
| *cia factbook* | tuple | country area, borders, population, *etc.* |

Table 5: List of sources used in the experiments and data fields extracted from them. Source type refers to how much data a source returns in response to a query — a single tuple or a list of tuples. For *airport* source, the type changed from a single tuple to a list over time.

result in the following precision, recall and accuracy values:

$$P = \frac{true\ positives}{true\ positives + false\ positives} = 0.73\,,$$

$$R = \frac{true\ positives}{true\ positives + false\ negatives} = 0.95\,,$$

$$A = \frac{true\ positives + true\ negatives}{positives\ + negatives} = 0.97\,.$$

These results are an improvement over those reported in (Lerman & Minton, 2000), which produced $P = 0.47, R = 0.95, A = 0.91$. The poor precision value reported in that work was due to 40 false positives obtained on the same data set. We attribute the improvements both to eliminating the specificity bias in the patterns learned by DataProG and to changing the feature set to include only the starting patterns and additional numeric features. Note that this improvement does not result simply from adding numeric features. To check this, we ran the verification experiments on a subset of data (the last 278 comparisons) using only the global numeric features and obtained $P = 0.92$ and $R = 0.55$, whereas using both patterns and numeric features results in values of $P = 0.71$ and $R = 1.00$ for the same data set.

### 3.1.2 Discussion of Results

Though we have succeeded in significantly reducing the number of false positives, we have not managed to eliminate them altogether. There are a number of reasons for their presence, some of which point to limitations in our approach.

We can split the types of errors into roughly three not entirely independent classes: improper tokenization, incomplete data coverage, and data format changes. The URL field (Table 6) accounted for a significant fraction of the false positives, in large part due to the design of our tokenizer, which splits text strings on punctuation marks. If the URL contains embedded punctuation (as part of the alphanumeric key associated with the user or session id), it will be split into a varying number of tokens, so that it is hard to capture the regularity of the field. The solution is to rewrite the tokenizer to recognize URLs for which well defined specifications exist. We will address this problem in our ongoing work. Our algorithm also failed sometimes (*e.g.*, *arrowlist*, *showtimes*) when it learned very long and specific descriptions. It is worth pointing out, however, that it performed correctly in over two dozen comparisons for these sources. These types of errors are caused by incomplete data coverage: a larger, more varied training data set would produce more general patterns, which would perform better on the verification task. A striking example of the data coverage problem occurred for the stock quotes source: the day the training data was collected, there were many more down movements in the stock price than up, and the opposite was true on the day the test data was collected. As a result, the price change fields for those two days were dissimilar. Finally, because DataProG learns the format of data, false positives will inevitably result from changes in the data format and do not indicate a problem with the algorithm. This is the case for the factbook source, where the units of area changed from "km2" to "sq km".

**hotel, mapquest** (5 cases): URL field contains alphanumeric keys, with embedded punctuation symbols. The tokenizer splits the field into many tokens. The key or its format changes from:

> http://…&Stamp=Q4aaiEGSp68*itn/hot%3da11204,itn/agencies/newitn… to
> http://…&Stamp=8∼bEgGEQrCo*itn/hot%3da11204,itn/agencies/newitn…
> On one occasion, the server name inside the URL changed: from
> http://enterprise.mapquest.com/mqmapgend?MQMapGenRequest=… to
> http://sitemap.mapquest.com/mqmapgend?MQMapGenRequest=…

**showtimes, arrowlist** (5 cases ): Instance of the showtimes field and part number and description fields (arrowlist) are very long. Many long, overly specific patterns are learned for these fields: *e.g.*,
< ( NUMBER : 2DIGIT ALLCAPS ) , ( SMALL : 2DIGIT ) , ( SMALL : 2DIGIT ) , ( 4 : 2DIGIT ) , 6 : 2DIGIT , 7 : 2DIGIT , 9 : 2DIGIT , 10 : 2DIGIT >

**altavista** (1 case): Database of the search engine appears to have been updated. A different set of results is returned for each query.

**quote** (1 case): Data changed — there were many more positive than negative price movements in the test examples

**factbook** (1 case): Data format changed:
*from* <NUMBER km2 >
*to* <NUMBER sq km >

Table 6: List of sources of false positive results on the verification task

## 3.2 Wrapper Reinduction

If the wrapper stops extracting correctly, the next challenge is to rebuild it automatically (Cohen, 1999). The extraction rules for our wrappers (Muslea et al., 2001), as well as many others (cf. (Kushmerick et al., 1997; Hsu & Dung, 1998)), are generated by a machine learning algorithm, which takes as input several pages from a source and labeled examples of data to extract from each page. It is assumed that the user labeled all examples correctly. If we label at least a few pages for which the wrapper fails by correctly identifying examples of data on them, we can use these examples as input to the induction algorithm, such as STALKER,[5] to generate new extraction rules.[6] Note that we do not need to identify the data on every page — depending on how regular the data layout is, STALKER can learn extraction rules using a small number of correctly labeled pages. Our solution is to bootstrap the wrapper induction process (which learns landmark-based rules) by learning content-based rules. We want to re-learn the landmark-based rules, because for the types of sites we use, these rules tend to be much more accurate and efficient than content-based rules.

We employ a method that takes a set of training examples, extracted from the source when the wrapper was known to be working correctly, and a set of pages from the same source, and uses a mixture of supervised and unsupervised learning techniques to identify examples of the data field on new pages. We assume that the format of the data did not change. Patterns learned by DataProG play a significant role in the reinduction task. In addition to patterns, other features, such as the length of the training examples and structural information about pages are used. In fact, because page structure is used during a critical step of the algorithm, we discuss our approach to learning it in detail in the next paragraph.

### 3.2.1 Page Template Algorithm

Many Web sources use templates, or page skeletons, to automatically generate pages and fill them with results of a database query. This is evident in the example in Fig. 6. The template consists of the heading "RESULTS", followed by the number of results that match the query, the phrase "Click links associated with businesses for more information," then the heading "ALL LISTINGS," followed by the anchors "map," "driving directions," "add to My Directory" and the bolded phrase "Appears in the Category." Obviously, data is not part of the template — rather, it appears in the *slots* between template elements.

Given two or more example pages from the same source, we can induce the template used to generate them (Table 7). The template finding algorithm looks for all sequences of tokens — both HTML tags and text — that appear exactly once on each page. The algorithm works in the following way: we pick the smallest page in the set as the template seed. Starting with the first token on this page, we grow a sequence by appending tokens

---

5. It does not matter, in fact, matter which wrapper induction system is used. We can easily replace STALKER with HLRT (Kushmerick et al., 1997) to generate extraction rules.

6. In this paper we will only discuss wrapper reinduction for information sources that return a single tuple of results per page, or a detail page. In order to create data extraction rules for sources that return lists of tuples, the STALKER wrapper induction algorithm requires user to specify the first and last elements of the list, as well as at least two consecutive elements. Therefore, we need to be able to identify these data elements with a high degree of certainty.

(a)

(b)

Figure 6: Fragments of two Web pages from the same source displaying restaurant information.

to it, subject to the condition that the sequence appears on every page. If we managed to build a sequence that's at least three tokens long[7], and this sequence appears exactly once on each page, it becomes part of the page template. Templates play an important role in helping identify correct data examples on pages.

input:
      $P$ = set of $N$ Web pages
output:
      $T$ = page template
begin
      $p$ = shortest($P$)
      $T$ = null
      $s$ = null
      for $t$ = firsttoken($p$) to lasttoken($p$)
          $s'$ = concat($s, t$)
          if ( $s'$ appears on every page in $P$ )
            $s = s'$
            continue
          else
            $n = \sum_{page=1}^{N} \text{count}(s, \text{page})$
            if ( $n = N$ AND length($s$) $\geq 3$ )
              add-to-template($T, s$)
            end if
            $s$ = null
          end if
      end for
end

Table 7: Pseudocode of the template finding algorithm

### 3.2.2 Automatic Labeling Algorithm

Figure 7 is a schematic outline of the reinduction algorithm, which consists of automatic data labeling and wrapper induction. Because the latter aspect is described in detail in other work (Muslea et al., 2001), we focus the discussion below on the automatic data labeling algorithm.

First, DataProG learns the starting and ending patterns that describe the set of training examples. These training examples have been collected during wrapper's normal operation, while it was correctly extracting data from the Web source. The patterns are used to identify possible examples of the data field on the new pages. In addition to patterns, we also calculate the mean (and its variance) of the number-of-tokens in the training examples. Each new page is then scanned to identify all text segments that begin with one of the starting patterns and end with one of the ending patterns. Text segments that con-tain significantly more or fewer tokens than expected based on the old number-of-tokens distribution, are eliminated from the set of candidate extracts. The learned patterns are often too general and will match many, possibly hundreds, text segments on each page. Among these spurious text segments is the correct example of the data field. The rest of the discussion is concerned with identifying the correct examples of data on pages.

---

7. The best value for the minimum length for the page template element was determined empirically to be three.

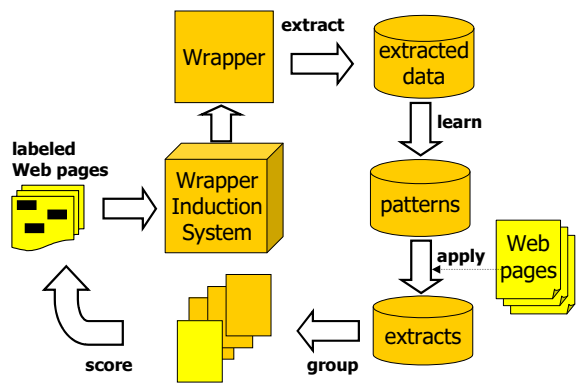

Figure 7: Schematic outline of the reinduction algorithm

We exploit some simple *a priori* assumptions about the structure of Web pages to help us separate interesting extracts from noise. We expect examples of the same data field to appear roughly in the same position and in the same context on each page. For example, Fig. 6 shows fragments of two Web pages from the same source displaying restaurant information. On both pages the relevant information about the restaurant appears after the heading "ALL LISTINGS" and before the phrase "Appears in the Category:". Thus, we expect the same field, *e.g.*, address, to appear in the *same place*, or slot, within the page template. Moreover, the information we are trying to extract will not usually be part of the page template; therefore, candidate extracts that are part of the page template can be eliminated from consideration. Restaurant address always follows restaurant name (in bold) and precedes the city and zip code, *i.e.*, it appears in the *same context* on every page. A given field is either visible to the user on every page, or it is invisible (part of an HTML tag) on every page. In order to use this information to separate extracts, we describe each candidate extract by a feature vector, which includes positional information, defined by the (page template) slot number and context. The context is captured by the adjacent tokens: one token immediately preceding the candidate extract and one token immediately following it. We also use a binary feature which has the value one if the token is visible to the user, and zero if it is part of an HTML tag. Once the candidate extracts have been assigned feature vectors, we split them into groups, so that within each group, the candidate extracts are described by the same feature vector.

The next step is to score groups based on their similarity to the training examples. We expect the highest scoring group to contain correct examples of the data field. One scoring method involves assigning a rank to the groups based on how many extracts they have in common with the training examples. This technique generally works well, because at least some of the data usually remains the same when the Web page layout changes. Of course, this assumption does not apply to data that changes frequently, such as weather information, flight arrival times, stock quotes, *etc.* However, we have found that even in these sources, there is enough overlap in the data that our approach works. If the scoring algorithm assigns zero to all groups, *i.e.*, there exist no extracts in common with the training examples, a second scoring algorithm is invoked. This scoring method follows the wrapper verification procedure and finds the group that is most similar to the training examples based on the patterns learned from the training examples.

The final step of the wrapper reinduction process is to provide the extracts in the top ranking group to the STALKER wrapper induction algorithm (Muslea et al., 2001) along with the new pages. STALKER learns data extraction rules for the changed pages. Note that examples provided to STALKER are required to be the correct examples of the field. If the set of automatically labeled examples includes false positives, STALKER will not learn correct extraction rules for that field. False negatives are not a problem, however. If the reinduction algorithm could not find the correct example of data on a page, that page is simply not used in the wrapper induction stage.

### 3.2.3 RESULTS

To evaluate the reinduction algorithm we used only the ten sources (listed in Table 5) that returned a single tuple of results per page, a detail page.[8] The method of data collection was described in Sec. 3.1.1. Over the period between October 1999 and March 2000 there were eight format changes in these sources. Since this set is much too small for evaluation purposes, we created an artificial test set by considering all ten data sets collected for each source during this period. We evaluated the algorithm by using it to extract data from Web pages for which correct output is known. Specifically, we took ten tuples from a set collected on one date, and used this information to extract data from ten pages (randomly chosen) collected at a later date, regardless of whether the source had actually changed or not. We reserved the remaining pages collected at a later date for testing the learned STALKER rules.

The output of the reinduction algorithm is a list of tuples extracted from ten pages, as well as extraction rules generated by STALKER for these pages. Though in most cases we were not able to extract every field on every pages, we can still learn good extraction rules with STALKER as long as few examples of each field are correctly labeled. We evaluated the reinduction algorithm in two stages: first, we checked how many data fields for each source were identified successfully; second, we checked the quality of the learned STALKER rules by using them to extract data from test pages.

**Extracting with content-based rules** We judged a data field to be successfully extracted if the automatic labeling algorithm was able to identify it correctly on at least two of the ten pages. This is the minimum number of examples STALKER needs to create extraction rules. In practice, such a low success rate only occurred for one field each in two of the sources: *Quote* and *Yahoo Quote*. For all other sources, if a field was successfully extracted, it was correctly identified in at least three, and in most cases almost all, of the pages in the set. A false positive occurred when the reinduction algorithm incorrectly identified some text on a page as a correct example of a data field. In many cases, false positives consisted of partial fields, *e.g.*, "Cloudy" rather than "Mostly Cloudy" (*yahoo weather*). A false negative occurred when the algorithm did not identify any examples of a data field. We ran the reinduction experiment attempting to extract the fields listed in Table 8. The second column of the table lists the fractions of data sets for which the field was successfully extracted. We were able to correctly identify fields 277 times across all data sets making 61 mistakes, of which 31 were attributed to false positives and 30 to the false negatives.

---

8. We did not use the *geocoder* and *cia factbook* wrappers in the experiments. The *geocoder* wrapper accessed the source through another application; therefore, the pages were not available to us for analysis. The reason for excluding the *factbook* is that it is a plain text source, while our methods apply to Web pages. Note also that in the verification experiments, we had two wrappers for the *mapquest* source, each extracting different data. In the experiments described below, we used the one that contained more data for this time period.

There are several reasons the reinduction algorithm failed to operate perfectly. In many cases the reason was the small training set.[9] We can achieve better learning for the yellow-pages-type sources *Bigbook* and *Smartpages* by using more training examples (see Fig. 8). In two cases, the errors were attributable to changes in the format of data, which resulted in the failure of patterns to capture the structure of data correctly: *e.g.*, the *airport* source changed airport names from capitalized words to allcaps, and in the *Quote* source in which the patterns were not able to identify negative price changes because they were learned for a data set in which most of the stocks had a positive price change. For two sources the reinduction algorithm could not distinguish between correct examples of the field and other examples of the same data type: for the *Quote* source, in some cases it extracted opening price or high price for the stock price field, while for the *yahoo weather* source, it extracted high or low temperature, rather than the current temperature. This problem was also evident in the *Smartpages* source, where the city name appeared in several places on the page. In these cases, user intervention or meta-analysis of the fields may be necessary to improve results of data extraction.

**Extracting with landmark-based rules**  The final validation experiment consisted of using the automatically generated wrappers to extract data from test pages. The last three columns in Table 8 list precision, recall and accuracy for extracting data from test pages. The performance is very good for most fields, with the notable exception of the *STATE* field of *Bigbook* source. For that field, the pattern <ALLCAPS> was overly general, and a wrong group received the highest score during the scoring step of the reinduction algorithm. The average precision and recall values were $P = 0.90$ and $R = 0.80$.

Within the data set we studied, five sources, listed in Table 9, experienced a total of seven changes. In addition to these sources, the *airport* source changed the format of the data it returned, but since it simultaneously changed the presentation of data from a detail page to a list, we could not use this data to learn STALKER rules. Table 9 shows the performance of the automatically reinduced wrappers for the changed sources. For most fields precision $P$, the more important of the performance measures, is close to its maximum value, indicating that there were few false positives. However, small values of recall indicate that not all examples of these fields were extracted. This result can be traced to a limitation of our approach: if the same field appears in a different context, more than one rule is necessary to extract it from a source. In such cases, we extract only a subset of the examples that share the same context, but ignore the rest of the examples.

---

9. Limitations in the data collection procedure prevented us from accumulating large data sets for all sources; therefore, in order to keep the methodology uniform across all sources, we decided to use smaller training sets.

| source/field | ex % | p | r |
|---|---|---|---|
| *airport* code | 100 | 1.0 | 1.0 |
| *airport* name | 90 | 1.0 | 1.0 |
| *Amazon* author | 100 | 97.3 | 0.92 |
| *Amazon* title | 70 | 98.8 | 0.81 |
| *Amazon* price | 100 | 1.0 | 0.99 |
| *Amazon* ISBN | 100 | 1.0 | 0.91 |
| *Amazon* availability | 60 | 1.0 | 0.86 |
| *Barnes&Noble* author | 100 | 0.93 | 0.96 |
| *Barnes&Noble* title | 80 | 0.96 | 0.62 |
| *Barnes&Noble* price | 90 | 1.0 | 0.68 |
| *Barnes&Noble* ISBN | 100 | 1.0 | 0.95 |
| *Barnes&Noble* availability | 90 | 1.0 | 0.92 |
| *Bigbook* name | 70 | 1.0 | 0.76 |
| *Bigbook* street | 90 | 1.0 | 0.87 |
| *Bigbook* city | 70 | 0.91 | 0.98 |
| *Bigbook* state | 100 | 0.04 | 0.50 |
| *Bigbook* phone | 90 | 1.0 | 0.30 |
| *mapquest* time | 100 | 1.0 | 0.98 |
| *mapquest* distance | 100 | 1.0 | 0.98 |
| *Quote* pricechange | 50 | 0.38 | 0.36 |
| *Quote* ticker | 63 | 0.93 | 0.87 |
| *Quote* volume | 100 | 1.0 | 0.88 |
| *Quote* shareprice | 38 | 0.46 | 0.60 |
| *Smartpages* name | 80 | 1.0 | 0.82 |
| *Smartpages* street | 80 | 1.0 | 0.52 |
| *Smartpages* city | 0 | 0.68 | 0.58 |
| *Smartpages* state | 100 | 1.0 | 0.70 |
| *Smartpages* phone | 100 | 0.99 | 1.0 |
| *Yahoo Quote* pricechange | 100 | 1.0 | 0.41 |
| *Yahoo Quote* ticker | 100 | 1.0 | 0.98 |
| *Yahoo Quote* volume | 100 | 1.0 | 0.99 |
| *Yahoo Quote* shareprice | 80 | 1.0 | 0.59 |
| *Washington Post* price | 100 | 1.0 | 1.0 |
| *Weather* temp | 40 | 0.36 | 0.82 |
| *Weather* outlook | 90 | 0.83 | 1.0 |
| average | 83 | 0.90 | 0.80 |

Table 8: Reinduction results on ten Web sources. The first column lists the fraction of the fields for each source that were correctly extracted by the pattern-based algorithm. We judged the field to be extracted if the algorithm correctly identified at least two examples of it. The last two columns list precision and recall on the data extraction task using the reinduced wrappers.

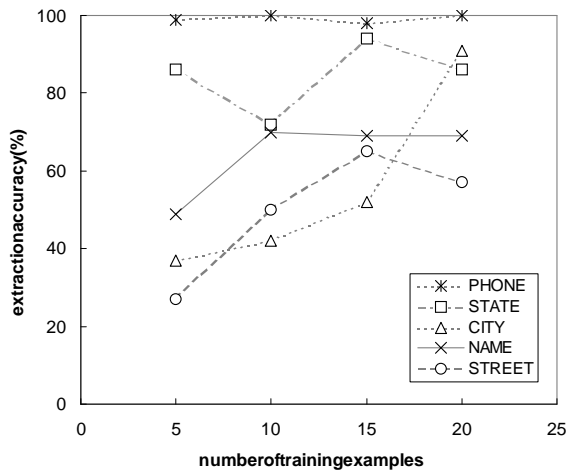

Figure 8: Performance of the reinduction algorithm for the fields in the *Smartpages* source as the size of the training set is increased

| source/field | P | R | A | source/field | P | R | A |
|---|---|---|---|---|---|---|---|
| *Amazon* author | 1.0 | 1.0 | 1.0 | *Smartpages* name | 1.0 | 0.9 | 0.9 |
| *Amazon* title | 1.0 | 0.7 | 0.7 | *Smartpages* street | N/A | 0.0 | 0.0 |
| *Amazon* price | 0.9 | 0.9 | 0.9 | *Smartpages* city | 0.0 | 0.0 | 0.0 |
| *Amazon* ISBN | 1.0 | 0.9 | 0.9 | *Smartpages* state | 1.0 | 0.9 | 0.9 |
| *Amazon* availability | 1.0 | 0.9 | 0.9 | *Smartpages* phone | N/A | 0.0 | 0.0 |
| *Barnes&Noble* author | 1.0 | 0.5 | 0.5 | *Yahoo Quote* pricechange | 1.0 | 0.2 | 0.2 |
| *Barnes&Noble* title | 1.0 | 0.8 | 0.8 | *Yahoo Quote* ticker | 1.0 | 0.5 | 0.5 |
| *Barnes&Noble* price | 1.0 | 1.0 | 1.0 | *Yahoo Quote* volume | 1.0 | 0.7 | 0.7 |
| *Barnes&Noble* ISBN | 1.0 | 1.0 | 1.0 | *Yahoo Quote* shareprice | 1.0 | 0.7 | 0.7 |
| *Barnes&Noble* availability | 1.0 | 1.0 | 1.0 | | | | |
| *Quote* pricechange | 0.0 | 0.0 | 0.0 | *Quote* volume | 1.0 | 1.0 | 1.0 |
| *Quote* ticker | 1.0 | 1.0 | 1.0 | *Quote* shareprice | 0.0 | N/A | 0.0 |

Table 9: Precision, recall, and accuracy of the learned STALKER rules for the changed sources

As mentioned earlier, we believe we can achieve better performance for the yellow-pages-type sources *Bigbook* and *Smartpages* by using more training examples. Figure 8 shows the effect increasing the size of the training example set on the performance of the automatically generated wrappers for the *Smartpages* source. As the number of training examples goes up, the accuracy of most extracted fields goes up.

### 3.2.4 LISTS

We have also applied the reinduction algorithm to extract data from pages containing lists of tuples, and, in many cases, have successfully extracted at least several examples of each field from several pages. However, in order to learn the correct extraction rules for sources returning lists of data, STALKER requires that the first, last and at least two consecutive list elements be correctly specified. The methods presented here cannot guarantee that the required list elements are extracted, unless all the list elements are extracted. We are currently working on new approaches to data extraction from lists (Lerman, Knoblock, & Minton, 2001) that will enable us to use STALKER to learn the correct data extraction rules.

## 4. Previous Work

There has been a significant amount of research activity in the area of pattern learning. In the section below we discuss two approaches, grammar induction and relational learning, and compare their performance to DATAPROG on tasks in the Web wrapper application domain. In Section 4.2 we review previous work on topics related to wrapper maintenance, and in Section 4.3 we discuss related work in information extraction and wrapper induction.

### 4.1 Pattern Learning

#### 4.1.1 GRAMMAR INDUCTION

Several researchers have addressed the problem of learning the structure, or patterns, describing text data. In particular, grammar induction algorithms have been used in the past to learn the common structure of a set of strings. Carrasco and Oncina proposed ALERGIA (Carrasco & Oncina, 1994), a stochastic grammar induction algorithm that learns a regular language from positive examples of the language. ALERGIA starts with a finite state automaton (FSA) that is initialized to be a prefix tree that represents all the strings of the language. ALERGIA uses a state-merging approach (Angluin, 1982; Stolcke & Omohundro, 1994) in which the FSA is generalized by merging pairs of statistically similar (at some significance level) subtrees. Similarity is based purely on the relative frequencies of substrings encoded in the subtrees. The end result is a minimum FSA that is consistent with the grammar.

Goan et al. (Goan et al., 1996) found that when applied to data domains commonly found on the Web, such as addresses, phone numbers, *etc.*, ALERGIA tended to merge too many states, resulting in an over-general grammar. They proposed modifications to ALERGIA, resulting in algorithm WIL, aimed at reducing the number of faulty merges. The modifications were motivated by the observation that each symbol in a string belong to one of the following syntactic categories: NUMBER, LOWER, UPPER and DELIM. When viewed on the syntactic level, data strings contain additional structural information that can be effectively exploited to reduce the number of faulty merges. WIL merges two subtrees if they are similar (in the ALERGIA sense) and also if, at every level, they contain nodes that are of the same syntactic type. WIL also adds a wildcard generalization step in which the transitions corresponding to symbols of the same category that are approximately evenly distributed over the range of that syntactic type (*e.g.*, 0–9 for numerals) are replaced with a single transition corresponding to the type (*e.g.*, NUMBER). Goan *et al.* demonstrated that the grammars learned by WIL were more effective in recognizing new strings in several relevant Web domains.

We compared the performance of WIL to DataProG on the wrapper verification task. We used WIL to learn the grammar on the token level using data examples extracted by the wrappers, not on the character level as was done by Goan *et al.*Another difference from Goan *et al.* was that, whereas they needed on the order of 100 strings to arrive at a high accuracy rate, we have on the order of 20–30 examples to work with. Note that we can no longer apply the wildcard generalization step to the FSA because we would need many more examples to decide whether the token is approximately evenly distributed over that syntactic type. Instead, we compare DataProG against two versions of WIL: one without wildcard generalization (WIL1), and one in which every token in the initial FSA is replaced by its syntactic type (WIL2). In addition to the syntactic types used by Goan *et al.*, we also had to introduce another type ALNUM to be consistent with the patterns learned by DataProG. Neither version of WIL allows for multi-level generalization.

The algorithms were tested on data extracted by wrappers from 26 Web sources on ten different occasions over a period of several months (see Sec. 3.1). Results of 20–30 queries were stored every time. For each wrapper, one data set was used as the training examples, and the data set extracted on the very next date was used as test examples. We used WIL1 and WIL2 to learn the grammar of each field of the training examples and then used the grammar to recognize the test examples. If the grammar recognized more than 80% of the test examples of a data field, we concluded that it recognized the entire data field; otherwise, we concluded that the grammar did not recognize the field, possibly because the data itself has changed. This is the same procedure we used in the wrapper verification experiments, and it is described in greater detail in Section 3.1.1. Over the period of time covered by the data, there were 21 occasions on which a Web site changed, thereby causing the data extracted by the wrapper to change as well. The precision and recall values for WIL1 (grammar induction on specific tokens) were $P = 0.20$, and $R = 0.81$; for WIL2 (grammar induction on wildcards representing tokens' syntactic categories) the values were $P = 0.55$ and $R = 0.76$. WIL1 learned an overly specific grammar, which resulted in a high rate of false positives on the verification task, while WIL2 learned an overly general grammar, resulting in slightly more false negatives. The recall and precision value of DataProG for the same data were $P = 0.73$ and $R = 1.0$.

Recently Thollard *et al.* (Thollard, Dupont, & de la Higuera, 2000) introduced MDI, an extension to ALERGIA. MDI has been shown to generate better grammars in at least one domain by reducing the number of faulty merges between states . MDI replaces ALERGIA's state merging criterion with a more global measure that attempts to minimize the Kullback-Leibler divergence between the learned automaton and the training sample while at the same time keeping the size of the automaton as small as possible. It is not clear whether MDI (or a combination of MDI/WIL) will lead to better grammars for common Web data types. We suspect not, because regular grammars capture just a few of the multitude of data types found on the Web. For example, business names, such as restaurant names shown in Table 2 may not have a well defined structure, yet many of them start with two capitalized words and end with the word "Restaurant" — which constitute patterns learned by DataProG.

#### 4.1.2 RELATIONAL LEARNING

As a sequence of $n$ tokens, a pattern can also be viewed as a non-recursive $n$-ary predicate. Therefore, we can use a relation-learning algorithm like FOIL (Quinlan, 1990) to learn them. Given a set of positive and negative examples of a class, FOIL learns first order predicate logic clauses defining the class. Specifically, it finds a discriminating description that covers many positive and none of the negative examples.

We used Foil.6 with the no-negative-literals option to learn patterns describing several different data fields. In all cases the closed world assumption was used to construct negative examples from the known objects: thus, for the *Bigbook* source, names and addresses were the negative examples for the phone number class. We used the following encoding to translate the training examples to allow foil.6 to learn logical relations. For each data field, FOIL learned clauses of the form

$$data\ field(A) := P(A)\ followed\_by(A, B)\ P(B)\,, \tag{3}$$

as a definition of the field, where $A$ and $B$ are tokens, and the terms on the right hand side are predicates. The predicate $followed\_by(A, B)$ expresses the sequential relation between the tokens. The predicate $P(A)$ allows us to specify the token $A$ as a specific token (*e.g.*, $John(A)$) or a general type (*e.g.*, UPPER($A$), ALPHA($A$)), thus, allowing FOIL the same multi-level generalization capability as DATAPROG.

We ran Foil.6 on the examples associated with the *Bigbook* (see Tables 2–3). The relational definitions learned by Foil.6 from these examples are shown in Table 10.

In many cases, there were similarities between the definitions learned by FOIL and the patterns learned by DATAPROG, though clauses learned by FOIL tended to be overly general. Another problem was when given examples of a class with little structure, such as names and book titles, FOIL tended to create clauses that covered single examples, or it failed to find any clauses. In general, the description learned by FOIL depended critically on what we supplied as negative examples of that field. For example, if we were trying to learn a definition for book titles in the presence of prices, FOIL would learn that something that starts with a capitalized word is a title. If author names were supplied as negative examples as well, the learned definition would have been different. Therefore, using FOIL in situations where the complete set of negative examples is not known or available, is problematic.

### 4.2 Wrapper Maintenance

Kushmerick (Kushmerick, 1999) addressed the problem of wrapper verification by proposing an algorithm RAPTURE to verify that a wrapper correctly extracts data from a Web page. In that work, each data field was described by a collection of global features, such as word count, average word length, and density of types, *i.e.*, proportion of characters in the training examples that are of an HTML, alphabetic, or numeric type. RAPTURE calculated the mean and variance of each feature's distribution over the training examples. Given a set of queries for which the wrapper output is known, RAPTURE generates a new result for each query and calculates the probability of generating the observed value for every feature. Individual feature probabilities are then combined to produce an overall probability that the wrapper has extracted data correctly. If this probability exceeds a certain threshold,

```
*** Warning:  the following definition does not cover 23 tuples in the relation
```
NAME(A) := ALLCAPS(A), followed_by(A,B)
NAME(A) := UPPER(A), followed_by(A,B), NUMBER(B)
NAME(A) := followed_by(A,B), Venice(B)

STREET(A) := LARGE(A), followed_by(A,B)
STREET(A) := MEDIUM(A), followed_by(A,B), ALPHANUM(B)

```
** Warning:  the following definition does not cover 9 tuples in the relation
```
CITY(A) := Los(A)
CITY(A) := Marina(A)
CITY(A) := New(A)
CITY(A) := Brooklyn(A)
CITY(A) := West(A), followed_by(A,B), ALPHA(B)

STATE(A) := CA(A)
STATE(A) := NY(A)

PHONE(A) := ((A)

Table 10: Definitions learned by foil.6 for the *Bigbook* source

RAPTURE decides that the wrapper is correct; otherwise, that it has failed. Kushmerick found that the HTML density alone can correctly identify almost all of the changes in the sources he monitored. In fact, adding other features in the probability calculation significantly reduced algorithm's performance. We compared RAPTURE's performance on the verification task to our approach, and found that RAPTURE missed 17 wrapper changes (false negatives) if it relied solely on the HTML density feature. [10]

There has been relatively little prior work on the wrapper reinduction problem. Cohen (Cohen, 1999) adapted WHIRL, a "soft" logic that incorporates a notion of statistical text similarity, to recognize page structure of a narrow class of pages: those containing simple lists and simple hotlists (defined as anchor-URL pairs). Previously extracted data, combined with page structure recognition heuristics, was used to reconstruct the wrapper once the page structure changed. Cohen conducted wrapper maintenance experiments using original data and corrupted data as examples for WHIRL. However, his procedure for corrupting data was neither realistic nor representative of how data on the Web changes. Although we cannot at present guarantee good performance of our algorithm on the wrapper reinduction for sources containing lists, we handle the realistic data changes in Web sources returning detail pages.

---

10. Although we use a different statistical test and cannot compare the performance of our algorithm to RAPTURE directly, we doubt that it would outperform our algorithm on our data set if it used all global numeric features, because, as we noted in Section 3.1.1, using patterns as well as global numeric features in the verification task outperforms using numeric features only.

### 4.3 Information Extraction

Our system, as used in the reinduction task, is related in spirit to the many information extraction (IE) systems developed both by our group and others in that it uses a learned representation of data to extract information from specific texts. Like wrapper induction systems (see (Muslea et al., 2001; Kushmerick et al., 1997; Freitag & Kushmerick, 2000)), it is domain independent and works best with semi-structured data, *e.g.*, Web pages. It does not handle free text as well as other systems, such as AutoSlog (Riloff, 1993) and Whisk (Soderland, 1999), because free text has fewer non-trivial regularities the algorithm can exploit. Unlike wrapper induction, it does not extract data based on the features that appear near it in text, but rather based on the content of data itself. However, unlike Whisk, which also learns content rules, our reinduction system represents each field independently of the other fields, which can be an advantage, for instance, when a web source changes the order in which data fields appear. Another difference is that our system is designed to run automatically, without requiring any user interaction to label informative examples. In the main part because it is purely automatic, the reinduction system fails to achieve the accuracy of other IE systems which rely on labeled examples to train the system; however, we do not see it as a major limitation, since it was designed to *complement* existing extraction tools, rather than supersede them. In other words, we consider the reinduction task to be successful if it can accurately extract a sufficient number of examples to use in a wrapper induction system. The system can then use the resulting wrapper to accurately extract the rest of the data from the source.

There are many similarities between our approach and that used by the RoadRunner system, developed concurrently with our system and reported recently in (Crescenzi, Mecca, & Merialdo, 2001b, 2001a). The goal of that system is to automatically extract data from Web sources by exploiting similarities in page structure across multiple pages. RoadRunner works by inducing the grammar of Web pages by comparing several pages containing long lists of data. The grammar is expressed at the HTML tag level, so it is similar to the extraction rules generated by Stalker. The RoadRunner system has been shown to successfully extract data from several Web sites. The two significant differences between that work and ours are (i) they do not have a way of detecting changes to know when the wrapper has to be rebuilt and (ii) our reinduction algorithm works on detail pages only, while RoadRunner works only on lists. We believe that our data-centric approach is more flexible and will allow us to extract data from more diverse information sources than the RoadRunner approach that only looks at page structure.

## 5. Conclusion

In this paper we have described the DataProG algorithm, which learns structural information about a data field from a set of examples of the field. We use these patterns in two Web wrapper maintenance applications: (i) **verification** — detecting when a wrapper stops extracting data correctly from a Web source, and (ii) **reinduction** — identifying new examples of the data field in order to rebuild the wrapper if it stops working. The verification algorithm performed with an accuracy of 97%, much better than results reported in our earlier work (Lerman & Minton, 2000). In the reinduction task, the patterns were used to identify a large number of data fields on Web pages, which were in turn used to automatically learn Stalker rules for these Web sources. The new extraction rules were validated by using them to successfully extract data from sets of test pages.

There remains work to be done on wrapper maintenance. Our current algorithms are not sufficient to automatically re-generate Stalker rules for sources that return lists of tuples. However, preliminary results indicate (Lerman et al., 2001) that it is feasible to combine information about the structure of data with *a priori* expectations about the structure of Web pages containing lists to automatically extract data from lists and assign it to rows and columns. We believe that these techniques will eventually eliminate the need for the user to mark up Web pages and enable us to automatically generate wrappers for Web sources. Another exciting direction for future work is using the DataProG algorithm to automatically create wrappers for new sources in some domain given existing wrappers for other sources in the same domain. For example, we can learn the author, title and price fields for the *AmazonBooks* source, and use them to extract the same fields on the *Barnes&NobleBooks* source. Preliminary results show that this is indeed feasible. Automatic wrapper generation is an important cornerstone of information-based applications, including Web agents.

## 6. Acknowledgments

We would like to thank Priyanka Pushkarna for carrying out the wrapper verification experiments.

The research reported here was supported in part by the Defense Advanced Research Projects Agency (DARPA) and Air Force Research Laboratory under contract/agreement numbers F30602-01-C-0197, F30602-00-1-0504, F30602-98-2-0109, in part by the Air Force Office of Scientific Research under grant number F49620-01-1-0053, in part by the Integrated Media Systems Center, a National Science Foundation (NSF) Engineering Research Center, cooperative agreement number EEC-9529152 and in part by the NSF under award number DMI-0090978. The U.S. Government is authorized to reproduce and distribute reports for Governmental purposes notwithstanding any copy right annotation thereon. The views and conclusions contained herein are those of the authors and should not be interpreted as necessarily representing the official policies or endorsements, either expressed or implied, of any of the above organizations or any person connected with them.

## References

Abramowitz, M., & Stegun, I. A. (1964). *Handbook of mathematical functions with formulas, graphs and mathematical tables*. Applied Math. Series 55. National Bureau of Standards, Washington, D.C.

Ambite, J.-L., Barish, G., Knoblock, C. A., Muslea, M., Oh, J., & Minton, S. (2002). Getting from here to there: Interactive planning and agent execution for optimizing travel. In *The Fourteenth Innovative Applications of Artificial Intelligence Conference (IAAI-2002), Edmonton, Alberta, Canada, 2002*.

Angluin, D. (1982). Inference of reversible languages. *Journal of the ACM, 29*(3), 741–765.

Brazma, A., Jonassen, I., Eidhammer, I., & Gilbert, D. (1995). Approaches to the automatic discovery of patterns in biosequences. Tech. rep., Department of Informatics, University of Bergen.

Carrasco, R. C., & Oncina, J. (1994). Learning stochastic regular grammars by means of a state merging method. *Lecture Notes in Computer Science*, *862*, 139.

Chalupsky, H., et al. (2001). Electric elves: Applying agent technology to support human organizations. In *Proceedings of the Thirteenth Annual Conference on Innovative Applications of Artificial Intelligence (IAAI-2001), Seattle, WA*.

Cohen, W. W. (1999). Recognizing structure in web pages using similarity queries. In *Proc. of the 16th National Conference on Artificial Intelligence (AAAI-1999)*, pp. 59–66.

Crescenzi, V., Mecca, G., & Merialdo, P. (2001a). Automatic web information extraction in the ROADRUNNER system. In *Proceedings of the International Workshop on Data Semantics in Web Information Systems (DASWIS-2001)*.

Crescenzi, V., Mecca, G., & Merialdo, P. (2001b). ROADRUNNER: Towards automatic data extraction from large web sites. In *Proceedings of the 27th Conference on Very Large Databases (VLDB)* Rome, Italy.

Dietterich, T., & Michalski, R. (1981). Inductive learning of structural descriptions.. *Artificial Intelligence*, *16*, 257–294.

Doorenbos, R. B., Etzioni, O., & Weld, D. S. (1997). A scalable comparison-shopping agent for the world-wide webs. In *Proceeding of the First International Confence on Autonomous Agents, Marina del Rey*.

Freitag, D., & Kushmerick, N. (2000). Boosted wrapper induction. In *Proceedings of the 7th Conference on Artificial Intelligence (AAAI-2000)*, pp. 577–583. AAAI Press, Menlo Park, CA.

Goan, T., Benson, N., & Etzioni, O. (1996). A grammar inference algorithm for the world wide web.. In *Proceedings of AAAI Spring Symposium on Machine Learning in Information Access, Stanford University, CA*.

Hsu, C.-N., & Dung, M.-T. (1998). Generating finite-state transducers for semi-structured data extraction from the web. *Journal of Information Systems*, *23*, 521–538.

Knoblock, C. A., Lerman, K., Minton, S., & Muslea, I. (2001a). Accurately and reliably extracting data from the web: A machine learning approach. *IEEE Data Engineering Bulletin*, *23*(4), 33–41.

Knoblock, C. A., Minton, S., Ambite, J. L., Muslea, M., Oh, J., , & Frank, M. (2001b). Mixed-initiative, multi-source information assistants. In *The Tenth International World Wide Web Conference (WWW10), Hong Kong*.

Kushmerick, N. (1999). Regression testing for wrapper maintenance.. In *Proceedings of the 14th National Conference on Artificial Intelligence (AAAI-1999)*.

Kushmerick, N., Weld, D. S., & Doorenbos, R. B. (1997). Wrapper induction for information extraction. In *Proceedings of the Intl. Joint Conference on Artificial Intelligence (IJCAI)*, pp. 729–737.

Lerman, K., Knoblock, C. A., & Minton, S. (2001). Automatic data extraction from lists and tables in web sources. In *Proceedings of the workshop on Advances in Text Extraction and Mining (IJCAI-2001)* Menlo Park. AAAI Press.

Lerman, K., & Minton, S. (2000). Learning the common structure of data. In *Proceedings of the 15th National Conference on Artificial Intelligence (AAAI-2000)* Menlo Park. AAAI Press.

Muggleton, S. (1991). Inductive logic programming. *New Generation Computing, 8*, 295–318.

Muslea, I., Minton, S., & Knoblock, C. (1998). Wrapper induction for semistructured web-based information sources.. In *Proceedings of the Conference on Automated Learning and Discovery (CONALD)*.

Muslea, I., Minton, S., & Knoblock, C. A. (2001). Hierarchical wrapper induction for semistructured information sources. *Autonomous Agents and Multi-Agent Systems, 4*, 93–114.

Papoulis, A. (1990). *Probability and Statistics*. Prentice Hall, Englewood Cliffs, NJ.

Quinlan, J. R. (1990). Learning logical definitions from relations.. *Machine Learning, 5*(3), 239–266.

Quinlan, J. R. (1993). *C4.5: Programs for Machine Learning*. Morgan Kaufmann, San Mateo, CA.

Riloff, E. (1993). Automatically constructing a dictionary for information extraction tasks. In *Proceedings of the 11th National Conference on Artificial Intelligence*, pp. 811–816 Menlo Park, CA, USA. AAAI Press.

Soderland, S. (1999). Learning information extraction rules for semi-structured and free text. *Machine Learning, 34*(1-3), 233–272.

Stolcke, A., & Omohundro, S. (1994). Inference of finite-state probabilistic grammars. In *Proceedings of the 2nd Int. Colloquium on Grammar Induction, (ICGI-94)*, pp. 106–118.

Thollard, F., Dupont, P., & de la Higuera, C. (2000). Probabilistic DFA inference using Kullback-Leibler divergence and minimality. In *Proceedings of the 17th International Conf. on Machine Learning*, pp. 975–982. Morgan Kaufmann, San Francisco, CA.